\documentclass{article} 
\usepackage{main,times}

\usepackage{amsmath,amsfonts,bm}

\def\eqref#1{equation~\ref{#1}}

\def\1{\bm{1}}

\DeclareMathAlphabet{\mathsfit}{\encodingdefault}{\sfdefault}{m}{sl}
\SetMathAlphabet{\mathsfit}{bold}{\encodingdefault}{\sfdefault}{bx}{n}

\usepackage{hyperref}
\usepackage{url}

\usepackage[utf8]{inputenc} 
\usepackage[T1]{fontenc}    
\usepackage{hyperref}       
\usepackage{url}            
\usepackage{booktabs}       
\usepackage{amsfonts}       
\usepackage{nicefrac}       
\usepackage{microtype}      

\usepackage{graphicx,epstopdf}
\DeclareGraphicsExtensions{.ps}
\usepackage{subcaption}
\usepackage{float}
\floatstyle{plaintop}
\restylefloat{table}
\usepackage{amsthm,amsmath}
\usepackage{algorithm,algpseudocode}
\usepackage{setspace}
\renewcommand{\headrulewidth}{0pt}

\newtheorem{definition}{Definition}
\theoremstyle{plain}
\newtheorem{theorem}{Theorem}

\title{Penetrating the Fog: the Path to Efficient CNN Models}

\author{Kun Wan, Boyuan Feng, Shu Yang \& Yufei Ding \\
Department of Computer Science\\
University of California, Santa Barbara\\
Santa Barbara, CA 93106, USA \\
\texttt{\{kun,boyuan,shuyang1995,yufeiding\}@cs.ucsb.edu} \\
}

\iclrfinalcopy 
\begin{document}

\maketitle

\begin{spacing}{0.9}
\begin{abstract}

With the increasing demand to deploy convolutional neural networks (CNNs) on mobile platforms, architecture designs with efficient sparse kernels (SKs) were proposed, which could save more parameters than the standard convolution while maintaining the accuracy. Despite the great potential, neural network design with SKs still require a lot of expert knowledge and take ample time. In this paper, we propose a~\textit{search scheme} that effectively reduce the SK design space based on three aspects:~\textit{composition},~\textit{performance} and~\textit{efficiency}. Further, we completely eliminate the model training from our search scheme. Instead, an easily measurable quantity,~\emph{information field}, is identified and used for predicting the model accuracy in the searching process. Additionally, we provide detailed~\textit{efficiency analysis} on the final designs found by our scheme. Experimental studies validate the idea of our scheme by showing: 1) existing SK designs can be rediscovered by our scheme, and 2) our scheme is able to find designs which are more efficient in parameters and computation with similar or higher accuracy.
\end{abstract}

\section{Introduction}
\label{sec:intro}


CNNs have achieved unprecedented success in visual recognition tasks. The development of mobile devices drives the increasing demand to deploy these deep networks on mobile platforms such as cell phones and self-driving cars. However, CNNs are usually resource-intensive, making them difficult to deploy on these memory-constrained and energy-limited platforms.

To enable the deployment, one intuitive idea is to reduce the model size. Model compression is the major research trend for it. Previously several techniques have been proposed, including pruning~\citep{lecun1990optimal}, quantization~\citep{soudry2014expectation} and low rank approximation~\citep{denton2014exploiting}. Though these approaches can can offer a reasonable parameter reduction with minor accuracy degradation, they suffer from the three drawbacks: 1) the irregular network structure after compression, which limits performance and throughput on GPU; 2) the increased training complexity due to the additional compression or re-training process; and 3) the heuristic compression ratios depending on networks, which cannot be precisely controlled.

Recently the SK approach was proposed to mitigate these problems by directly training networks using structural (large granularity) sparse convolutional kernels with fixed compression ratios. The idea of SK was originally proposed as different types of convolutional approach. Later researchers explore their usages in the context of CNNs by combining some of these SKs to save parameters/computation against the standard convolution. For example, MobileNets~\citep{howard2017mobilenets} realize 7x parameter savings with only 1\% accuracy loss by adopting the combination of two SKs, depthwise convolution~\citep{sifre2014rigid} and pointwise convoluiton~\citep{lin2013network}, to replace the standard convolution in their networks.

However, despite the great potential with SK approach to save parameters/computation while maintaining accuracy, it is still mysterious in the field regarding how to craft an SK design with such potential (i.e., effective SK design). Prior works like MobileNet~\citep{howard2017mobilenets} and Xception~\citep{chollet2016xception} just adopt simple combinations of existing SKs, and no one really points out the reasons why they choose such kind of design. Meanwhile, it has been a long-existing question in the field whether there is any other SK design that is more efficient than all state-of-the-art ones while also maintaining a similar accuracy with the standard convolution.

To answer this question, a native idea is to try all possible combinations and get the final accuracy for each of them. Unfortunately, the number of combination will grow exponentially with the number of kernels in a design, and thus it is infeasible to train each of them. Specifically, even if we limit the design space to four common types of SKs -- group convolution~\citep{krizhevsky2012imagenet}, depthwise convolution~\citep{sifre2014rigid}, pointwise convolution~\citep{lin2013network} and pointwise group convolution~\citep{zhang2017shufflenet} -- the total number of possible combinations would be $4^k$, given that $k$ is the number of SKs we allow to use in a design (note that each SK can appear more than once in a design). 

In this paper, we craft the effective SK design by efficiently eliminating poor candidates from the large design space. Specifically, we reduce the design space from three aspects: composition, performance and efficiency. First, observing that in normal CNNs it is quite common to have multiple blocks which contain repeated patterns such as layers or structures, we eliminate the design space by ignoring the combinations including repeated patterns. Second, realizing that removing designs with large accuracy degradation would significantly reduce the design space, we identify an easily measurable quantity named~\emph{information field} behind various SK designs, which is closely related to the model accuracy. We get rid of designs that lead to a smaller ~\emph{information field} compared to the standard convolution model. Last, in order to achieve a better parameter efficiency, we remove redundant SKs in a design if the same size of~\emph{information field} is already retained by other SKs in the design. With all aforementioned knowledge, we present a SK scheme that incorporates the final four different designs manually reduced from the original design space.

Additionally, in practice, researchers would also like to select the most parameter/computation efficient SK designs based on their needs, which drives the demand to study the efficiency for different SK designs. Previously no research has investigated on the efficiency for any SK design. In this paper, three aspects of efficiency are addressed for each of the SK designs in our scheme: 1) what are the factors which could affect the efficiency for each design? 2) how does each factor affect the efficiency alone? 3) when is the best efficiency achieved combining all these factors in different situations?

Besides, to verify the correctness of our idea we also study the relationship between our proposed scheme and existing designs. The comparisons show that all existing methods are either extreme cases or not optimal ones in terms of the parameter efficiency under our scheme. Further, we show that the accuracy of models composed of new designs in our scheme are better than that of all state-of-the-art methods under the same constraint of parameters, which implies that more efficient designs are constructed by our scheme and again validates the effectiveness of our idea.

The contributions of our paper can be summarized as follows:
\begin{itemize}
    \item We are the first in the field to point out that the~\emph{information field} is the key for the SK designs. Meanwhile we observe the model accuracy is positively correlated to the size of the~\emph{information field}.
    \item We present a SK scheme to illustrate how to eliminate the original design space from three aspects and incorporate the final 4 types of designs along with rigorous mathematical foundation on the efficiency.
    \item We discuss the connections between our proposed scheme and other existing designs like MobileNet~\citep{howard2017mobilenets}, Xception~\citep{chollet2016xception} and ShuffleNet~\citep{zhang2017shufflenet} and show that they are all specific instances of our proposed scheme.
    \item We provide some potential network designs which are in the scope of our scheme and have not been explored yet and show that they could have superior performances.
\end{itemize}

\section{Preliminaries}
\vspace{-0.1in}
We first give a brief introduction to the standard convolution and the four common types of SKs.
\vspace{-0.1in}
\subsection{Standard Convolution}
Standard convolution is the basic component in most CNN models, kernels of which can be described as a 4-dimensional tensor: $W\in \mathbb{R}^{C\times X\times Y\times F}$, where $C$ and $F$ are the numbers of the input and the output channels and $X$ and $Y$ are the spatial dimensions of the kernels. Let $I\in \mathbb{R}^{C\times U\times V}$ be the input tensor, where $U$ and $V$ denote the spatial dimensions of the feature maps. Therefore, the output activation at the output feature map $f$ and the spatial location $(x,y)$ can be expressed as,
\begin{equation*}
    T(f,x,y) = \sum_{c=1}^C\sum_{x'=1}^X\sum_{y'=1}^YI(c,x-x',y-y')W(c,x',y',f)
\end{equation*}

\subsection{Group Convolution}

Group convolution is first used in AlexNet~\citep{krizhevsky2012imagenet} for distributing the model over two GPUs. The idea of it is to split both input and output channels into disjoint groups and each output group is connected to a single input group and vice versa. By doing so, each output channel will only depend on a fraction of input channels instead of the entire ones, thus a large amount of parameters and computation could be saved. Considering the number of group as $M$, the output activation $(f, x, y)$ can be calculated as,

\begin{equation*}
    T(f,x,y) = \sum_{c'=1}^{C/M}\sum_{x'=1}^X\sum_{y'=1}^YI(\frac{C}{M}\lfloor\frac{f-1}{\frac{F}{M}}\rfloor+c',x-x',y-y')W(c',x',y',f)
\end{equation*}

\subsection{Depthwise Convolution}

The idea of depthwise convolution is similar to the group convolution, both of which sparsifies kernels in the channel extent. In fact, depthwise convolution can be regarded as an extreme case of group convolution when the number of groups is exactly the same with the number of input channels. Also notice that in practice usually the number of channels does not change after the depthwise convolution is applied. Thus, the equation above can be further rewritten as,

\begin{equation*}
    T(f,x,y) = \sum_{x'=1}^X\sum_{y'=1}^YI(f,x-x',y-y')W(x',y',f)
\end{equation*}

\subsection{Pointwise Convolution}

Pointwise convolution is actually a $1\times 1$ standard convolution. Different from the group convolution, pointwise convolution achieves the sparsity over the spatial extent by using kernels with $1\times 1$ spatial size. Similarly, the equation below shows how to calculate one output activation from the pointwise convolution in detail,

\begin{equation*}
    T(f,x,y) = \sum_{c=1}^CI(c,x,y)W(c,f)
\end{equation*}

\subsection{Pointwise Group Convolution}

To sparsify kernels in both the channel and the spatial extents, the group convolution can be combined together with the pointwise convolution, i.e., pointwise group convolution. Besides the use of $1\times 1$ spatial kernel size, in pointwise group convolution each output channel will also depend on a portion of input channels. The specific calculations for one output activation can be found from the equation below,

\begin{equation*}
    T(f,x,y) = \sum_{c'=1}^{C/M}I(\frac{C}{M}\lfloor\frac{f-1}{\frac{F}{M}}\rfloor+c',x,y)W(c',f)
\end{equation*}

\section{Sparse Kernel Scheme}
\label{sec:decompose}

Recall that the total number of combinations will grow exponentially with the number of kernels in a design, which could result in a large design space. In this paper, we craft the effective SK design (i.e., design that consumes less parameters but maintains accuracy with the standard convolution) by efficiently examining the design space.

Specifically, first we determine the initial design space by setting the maximum number of SKs (length). To decide this number, two aspects are considered: 1) in order to give the potential to find more efficient designs which have not been explored yet, the maximum length of SK design should be greater than the numbers of all state-of-the-art ones; 2) it is also obvious that the greater length is more likely to consume more parameters, which contradicts our goal to find more efficient designs. Therefore combining the two aspects together, we set the maximum length to 6, which is not only greater than the largest number (i.e., 3) in all current designs, but also makes designs with the maximum length could still be able to be more efficient than the standard convolution. 

\subsection{Reduce the Design Space}
We then start to reduce the design space from three aspects: composition, performance and efficiency. In the following paragraphs, we will introduce the three aspects in detail.
\paragraph{Composition.}

The overall layout in CNNs provides a good insight for us to quickly reduce the design space. Specifically, in normal CNNs it is quite common to have multiple stages/blocks which contain repeated patterns such as layers or structures. For example, in both VGG~\citep{simonyan2014very} and ResNet~\citep{he2016deep} there are 4 stages and inside each stage are several same repeated layers. Inspired by the fact, when we replace the standard convolution using various SK designs intuitively there is no need to add these repeated patterns to the original place of each standard convolutional layer. For example, suppose there are three types of SKs, A, B and C, then the following combinations should be removed as containing repeated patterns: AAAAAA, ABABAB and ABCABC. AAAAAA contains the repeated pattern of A, while ABABAB and ABCABC have the patterns of AB and ABC respectively.

Repeated patterns are also easy to detect, which makes the entire process extremely fast. To find such patterns, we can use the regular expression matching. The corresponding expression for the matched combinations should be $(.+?)\1+$, where $(.+?)$ denotes the first capturing group which contains at least one character, but as few as possible, and $\1+$ means try to match the same character(s) as most recently matched by the first group as many times as possible. As a result, we can efficiently eliminate the design space with the help of the regular expression.

\begin{figure}[!th]
\centering
\begin{subfigure}{0.2\linewidth}
  \centering
  \includegraphics[width=1.0\linewidth]{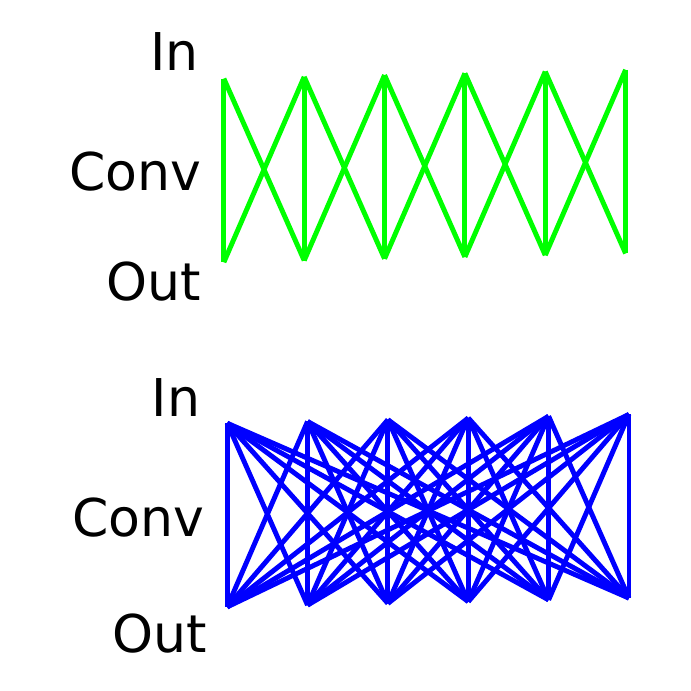}
  \caption{Standard}
  \label{fig:Standard}
\end{subfigure}%
\begin{subfigure}{0.2\linewidth}
  \centering
  \includegraphics[width=1.0\linewidth]{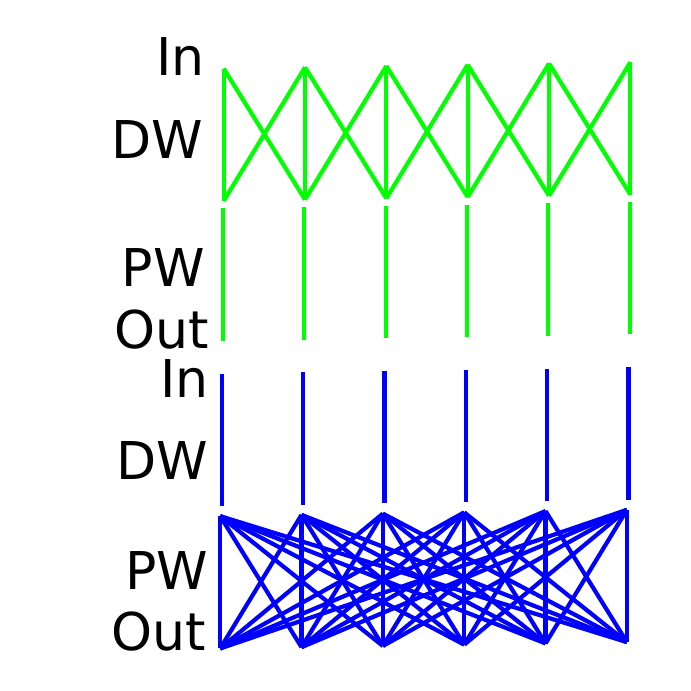}
  \caption{DW+PW}
  \label{fig:DW+PW}
\end{subfigure}%
\begin{subfigure}{0.2\linewidth}
  \centering
  \includegraphics[width=1.0\linewidth]{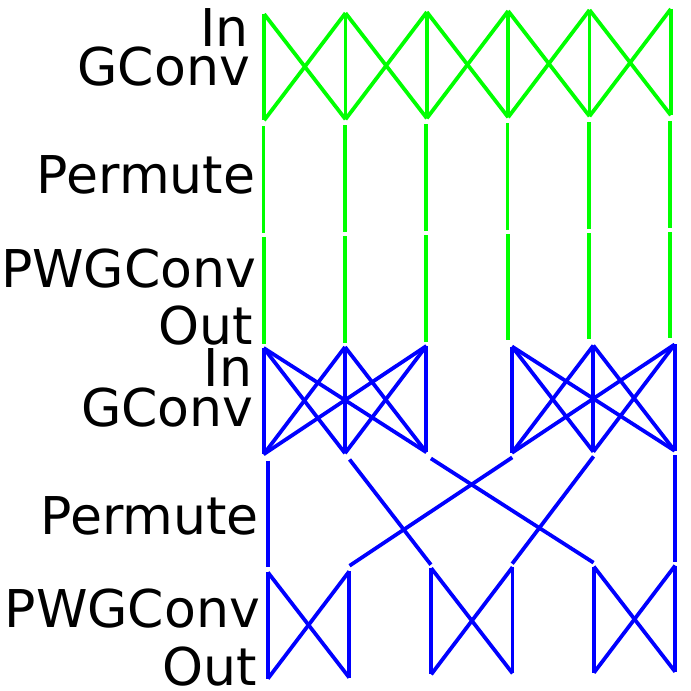}
  \caption{GC+PWG}
  \label{fig:GC+PWG}
\end{subfigure}%
\begin{subfigure}{0.2\linewidth}
  \centering
  \includegraphics[width=1.0\linewidth]{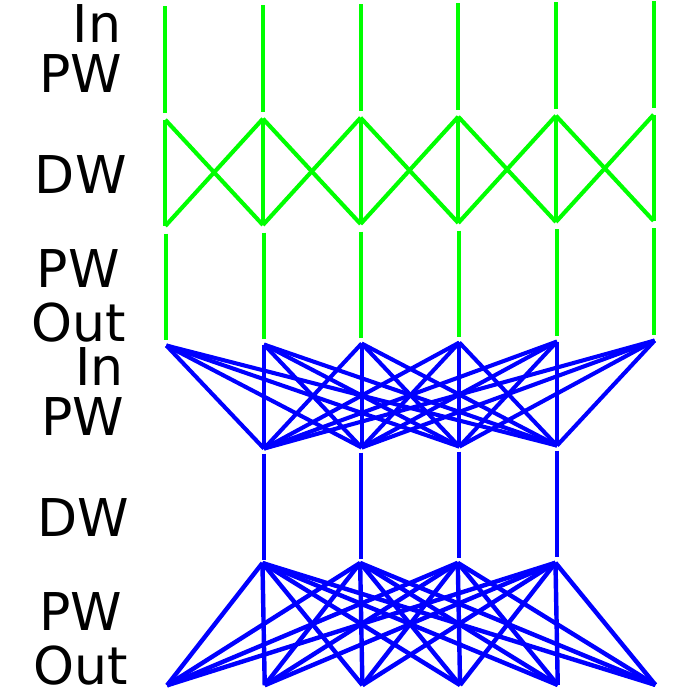}
  \caption{PW+DW+PW}
  \label{fig:PW+DW+PW}
\end{subfigure}%
\begin{subfigure}{0.2\linewidth}
  \centering
  \includegraphics[width=1.0\linewidth]{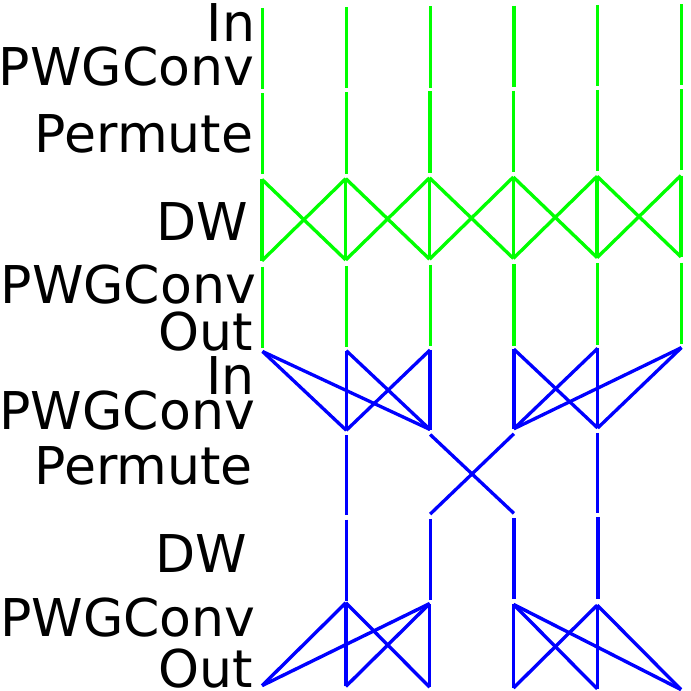}
  \caption{PWG+DW+PWG}
  \label{fig:PWG+DW+PWG}
\end{subfigure}
\caption{Spatial and channel dependency of the standard convolution and four different SK designs. The spatial kernel size is $3\times 3$. Green edges denote the spatial dependency of output activation and blue edges represent the channel dependency.}
\end{figure}

\paragraph{Performance.}

There are lots of SK designs that could result in large accuracy degradation, which gives us another opportunity to greatly reduce the design space. To get rid of them, we need an easily measurable (i.e., no training) property behind various designs that could directly indicate the final accuracy. Fortunately, after analyzing many prior works and conducting many experimental studies, we do find such property. We name it~\emph{information field}.

\begin{definition}(Information Field)
Information field is the area in input tensor which one or more convolutional layers use to generate one output activation. For one output tensor, sizes of information fields for all activations are usually the same.
\end{definition}


Figure~\ref{fig:Standard} shows the spatial and channel dependency for the standard convolution, from which we can also find out the size of~\emph{information field}. Assuming the spatial kernel size is $3\times 3$, starting from any output node in the figure we can see that in terms of the channel dimension each output channel will connect to all input channels and for the spatial dimensions one output activation will depend on activations inside a $3\times 3$ spatial area. Therefore the~\emph{information field} for the standard convolution will be (3, 3, C) where C is the number of input channels. 

We find that~\emph{information field} is the key behind all SK designs, and also observe the model accuracy is positively correlated to the size of~\emph{information field}, the idea of which is also validated by later experiments in Section~\ref{sec:study}.

With the help of~\emph{information field}, SK designs that would result in large accuracy degradation could be easily removed from the original design space without actually training the models. Specifically, first for each design we calculate the size of~\emph{information field} by adding up it sequentially from the leftmost kernel to the rightmost one. For example, we use a three-dimensional vector, (1,1,1), to represent the initial values of~\emph{information field} on three different dimensions (i.e., two spatial dimensions and one channel dimension), then corresponding values of the vector will be updated based on the known properties of the SK encountered. After the rightmost kernel, the final vector we get will be the size of~\emph{information field} for the design. Finally we compare it with that of the standard convolution. If the two sizes are the same, we will keep the design, otherwise we will simply discard it. For instance, the design composed of one depthwise convolution will be removed since the~\emph{information field} of it only contains one channel area instead of the full channel space from the standard convolution.

\paragraph{Efficiency.}

In order to achieve a better parameter and computation efficiency, we remove designs that include SKs that do not contribute to the~\emph{information field}. Specifically, there are two cases that could worsen the efficiency and should be regarded as inferior designs: 1) it can be easily verified that the size of~\emph{information field} will never decrease when passing through SKs in a design, thus there could be one situation that after one kernel, the size of~\emph{information field} still remains the same, which means the kernel does not help with regards to the~\emph{information field} even if the final size is the same as the standard convolution; 2) it is also possible that the same size of~\emph{information field} with the standard convolution is already retained by a fraction of SKs in a design, in which case, other kernels can also be considered as not contributing to the~\emph{information field}. In terms of efficiency designs in both of the two cases contain non-contributed kernels, therefore we can remove them from the original design space.




To effectively detect and delete such inferior designs within the two cases, we introduce an~\emph{early-stop mechanism} during the process to check the size of~\emph{information field} above. Specifically, as per the two cases we check two things when adding up~\emph{information field} from the leftmost kernel in a design: 1) we record the size of~\emph{information field} before entering each kernel and compare it with the new size calculated after that kernel. If the two sizes are the same, we immediately mark the current design as inferior; 2) we compare the new size of~\emph{information field} with that of the standard convolution. If the size is smaller, we will continue to add up~\emph{information field} from the next kernel, otherwise we will skip to the next design.



With all aforementioned knowledge, we manually reduce the original design space ($4^1+4^2+\cdots+4^6$) to 4 different types of SK designs\footnote{During the process to eliminate the design space, we allow channel permutation within the designs, and when a group convolution is encountered, we will try all possible numbers of groups to calculate the size of~\emph{information field}. As long as there is one group number that can pass the entire process, we will keep the design. In case there are multiple group numbers passing the process, we will consider them as same design.}. In the next section we will present the 4 final designs respectively. 

Also notice that other techniques to save parameters such as bottleneck structure~\citep{he2016deep} appear to be complimentary to our approach, which can be combined together to further improve parameter efficiency while maintaining accuracy. To validate this idea, we also consider the bottleneck structure when reducing the design space.

\subsection{Final Sparse Kernel Designs}

\paragraph{Depthwise Convolution + Pointwise Convolution.}
\label{sec:DW+PW}

Unlike the standard convolution which combines spatial and channel information together to calculate the output, the combination of depthwise convolution (DW) and pointwise convolution (PW) split the two kinds of information and deal with them separately. The output activation at location $(f,x,y)$ can be written as
\begin{equation*}
    T(f,x,y) = \sum_{c=1}^C[\sum_{x'=1}^X\sum_{y'=1}^YI(c,x-x',y-y')W_1(c,x',y')]W_2(c,f),
\end{equation*}
where $W_1$ and $W_2$ correspond to the kernels of depthwise convolution and pointwise convolution respectively. The dependency of such design is depicted in Figure~\ref{fig:DW+PW}, from which we can easily verify that the size of~\emph{information field} is the same with the standard convolution.

\paragraph{Group Convolution + Pointwise Group Convolution.}
\label{sec:GC+PWG}

The combination of group convolution (GC) and pointwise group convolution (PWG) can be regarded as an extension for the design above, where group convolution is applied on the pointwise convolution. However, simply using pointwise group convolution would reduce the size of~\emph{information field} on the channel dimension since depthwise convolution will not deal with any channel information. To recover the~\emph{information field} depthwise convolution is replaced with the group convolution. Meanwhile channel permutation should be added between the two layers. Assuming the number of channels does not change after the first group convolution, the output activation can be calculated as
\begin{equation*}
    T(f,x,y) = \sum_{k'=1}^{C/N}[\sum_{c'=1}^{C/M}\sum_{x'=1}^X\sum_{y'=1}^YI(\frac{C}{M}\lfloor\frac{k-1}{\frac{C}{M}}\rfloor+c',x-x',y-y')W_1(c',x',y',k)]W_2(k',f),
\end{equation*}
where $k = \frac{C}{N}\lfloor\frac{f-1}{\frac{F}{N}}\rfloor+k'$, $M$ and $N$ denote numbers of groups for group convolution and pointwise group convolution and $W_1$ and $W_2$ correspond to the kernels of group convolution and pointwise group convolution respectively. Figure~\ref{fig:GC+PWG} shows the~\emph{information field} of this design clearly.

\paragraph{Pointwise Convolution + Depthwise Convolution + Pointwise Convolution.}
\label{sec:PW+DW+PW}

Although two pointwise convolutions do not ensure a better efficiency in our scheme, the combination with bottleneck structure can help ease the problem, which makes it survive as one of the last designs. Following the normal practice we set bottleneck ratio to $1:4$, which implies the ratio of bottleneck channels to output channels. Also notice that more parameters could be saved if we place the depthwise convolution between the two pointwise convolutions since now depthwise convolution would only apply on a reduced number of channels. As a result, the output activation $T(f,x,y)$ is calculated as
\begin{equation*}
    T(f,x,y) = \sum_{k=1}^K[\sum_{x'=1}^X\sum_{y'=1}^Y[\sum_{c=1}^CI(c,x-x',y-y')W_1(c,k)]W_2(k,x',y')]W_3(k,f),
\end{equation*}
where $K$ denote the number of bottleneck channels and $W_1$, $W_2$ and $W_3$ correspond to the kernels of first pointwise convolution, depthwise convolution and second pointwise convolution respectively. Along with the equation Figure~\ref{fig:PW+DW+PW} shows that the~\emph{information field} of such design is same with the standard convolution.

\paragraph{Pointwise Group Convolution + Depthwise Convolution + Pointwise Group Convolution.}

The combination of two pointwise group convolutions and one depthwise convolution can also ensure the same size of~\emph{information field}. Similarly, channel permutation is needed. The bottleneck structure is also adopted to achieve a better efficiency. The output activation is calculated as
\begin{equation*}
    T(f,x,y) = \sum_{k'=1}^{K/N}[\sum_{x'=1}^X\sum_{y'=1}^Y[\sum_{c'=1}^{C/M}I(\frac{C}{M}\lfloor\frac{k-1}{\frac{K}{M}}\rfloor+c',x-x',y-y')W_1(c',k)]W_2(k,x',y')]W_3(k',f),
\end{equation*}
where $k = \frac{K}{N}\lfloor\frac{f-1}{\frac{F}{N}}\rfloor+k'$, $K$, $M$ and $N$ represent the number of bottleneck channels and numbers of groups for first pointwise group convolution and second pointwise group convolution and $W_1$, $W_2$ and $W_3$ correspond to the kernels of first pointwise group convolution, depthwise convolution and second pointwise group convolution respectively. Both the equation and Figure~\ref{fig:PWG+DW+PWG} could verify the same size of~\emph{information field} with the standard convolution.

\subsection{Efficiency Analysis}

In addition, we find that the efficiency for different designs in our scheme do not always overlap. Thus to save the pain for researchers to find the most parameter/computation efficient designs based on their needs, we study the efficiency for each of the designs. Specifically, we consider two real situations which are frequently encountered by researchers when applying SK designs (i.e., given the input and the output for a layer and given the total number of parameters for a layer) and give accurate conditions when the best efficiency could be achieved.






\subsubsection{Depthwise Convolution + Pointwise Convolution.}

\paragraph{Efficiency given the input and the output.}

Given the numbers of input and output channels $C$ and $F$. The total number of parameters after applying this design is $9C + CF$, and the number of parameters for standard convolution is $9CF$. Therefore the parameter efficiency of such method is $1/F + 1/9$ represented by the ratio of parameters after and before applying such design. Clearly, given $C$ and $F$, the parameter efficiency is always the same.

\paragraph{Efficiency given the total amount of parameters.}

It can be easily verified that given the total number of parameters the greatest width is reached when the best efficiency is achieved. Thus the condition for the best efficiency given the total amount of parameters should be the same with the one when the greatest width is reached.

The total number of parameters $P$ for the design can be expressed as
\begin{equation*}
    P = 3\cdot 3\cdot C + 1\cdot 1\cdot C \cdot F,
\end{equation*}
when studying the greatest width, we need to assume the ratio between $C$ and $F$ does not change, thus the number of output channels $F$ could be written like $F = \alpha \cdot C$ where usually $\alpha \in \mathbb{N}^+$. As a result, from the equation above when $P$ is fixed, the greatest width $G$ (i.e., $\frac{-9+\sqrt{81+4\alpha P}}{2\alpha}$) will also be fixed, which indicates that the parameter efficiency is always the same. 

\subsubsection{Group Convolution + Pointwise Group Convolution.}
\label{sec:proof}

\paragraph{Efficiency given the input and the output.}

Similarly, we use the ratio of parameters to show parameter efficiency of this design. Given $C$ and $F$, the number of parameters after using such design can be written as $3\cdot 3\cdot \frac{C}{M}\cdot C + 1\cdot 1\cdot \frac{C}{N}\cdot F = \frac{9C^2}{M} + \frac{CF}{N}$. Since the number of parameters for standard convolution is $9CF$, the ratio will become $\frac{C}{MF} + \frac{1}{9N}$. Notice that to ensure the same size of~\emph{information field} with standard convolution, in any input group of the second layer there should be at least one output channel from each one of the output groups of the first layer, therefore $M\cdot N$ should be less than or equal to the number of output channels from the first layer, i.e., $M\cdot N\leq C$. To further illustrate the relationship between the best parameter efficiency and the choices of $M$ and $N$, we have the following theorem (the proof is given in the Appendix):

\theoremstyle{plain}
\begin{theorem}
\label{the:eff}
With the same size of~\emph{information field}, the best parameter efficiency is achieved if and only if the product of the two group numbers equals the channel number of the intermediate layer.
\end{theorem}

As per the theorem, the best parameter efficiency can be achieved only when $M\cdot N = C$. Thus the ratio will become $\frac{N}{F} + \frac{1}{9N}$. When $F$ is a fixed number, $N$ is the only variable which could affect the efficiency. Since $\frac{N}{F} + \frac{1}{9N}\geq\frac{2}{3}\sqrt{\frac{1}{F}}$, the best efficiency can be achieved when $\frac{N}{F} = \frac{1}{9N}$, or $N = \frac{\sqrt{F}}{3}$.

\paragraph{Efficiency given the total amount of parameters.}

Given the total number of parameters $P$ for one design, both $M$ and $N$ could affect the width of the network. As per Theorem~\ref{the:eff} the greatest $C$ can be reached only when $C = M\cdot N$. When $F = \alpha \cdot C$, $P$ could be written like
\begin{align*}
    P & = 3\cdot 3\cdot N\cdot M\cdot N + 1\cdot 1\cdot M \cdot \alpha \cdot M\cdot N = MN(9N+\alpha M) \\
    & \geq MN\cdot 2\sqrt{9\alpha MN} = 6\sqrt{\alpha}C^{\frac{3}{2}}
\end{align*}
Given the number of parameters $P$, width C has a upper bound when $9N = \alpha M$, which is also the condition for the best efficiency. The greatest width $G$ is $(\frac{P}{6\sqrt{\alpha}})^{\frac{2}{3}}$.

\subsubsection{Pointwise Convolution + Depthwise Convolution + Pointwise Convolution.}

\paragraph{Efficiency given the input and the output.}

Same as before, given the number of input channels $C$, bottleneck channels $K$ and output channels $F$. After applying the design, the total amount of parameters is reduced to $1\cdot 1\cdot C\cdot K + 3\cdot 3\cdot K + 1\cdot 1\cdot K\cdot F = K(C + F + 9)$. The number of parameters for standard convolution is still $9CF$. Notice that $K = F/4$, therefore the ratio can be further expressed as $\frac{C + F + 9}{36C}$. Clearly, given $C$, $K$ and $F$, such design will also result in a fixed efficiency.

\paragraph{Efficiency given the total amount of parameters.}

When $F = \alpha \cdot C$ and $K = F/4$, the total number of parameters $P$ will be
\begin{equation*}
    P = 1\cdot 1\cdot C\cdot \frac{\alpha C}{4} + 3\cdot 3\cdot \frac{\alpha C}{4} + 1\cdot 1\cdot \frac{\alpha C}{4}\cdot \alpha C,
\end{equation*}
when $P$ is fixed, the greatest width $G$ is also fixed, i.e., $\frac{-9\alpha + \sqrt{81\alpha^2+16\alpha^2P+16\alpha P}}{2(\alpha^2+\alpha)}$.

\subsubsection{Pointwise Group Convolution + Depthwise Convolution + Pointwise Group Convolution}
\paragraph{Efficiency given the input and the output.}

We use the same way to evaluate parameter efficiency for this design. First, the number of parameters after applying such method is $1\cdot 1\cdot\frac{C}{M}\cdot K + 3\cdot 3\cdot K + 1\cdot 1\cdot\frac{K}{N} \cdot F = K(\frac{C}{M} + \frac{F}{N} + 9)$. The number for standard convolution is $9CF$. Since $K = F/4$ and as per Theorem~\ref{the:eff} the best parameter efficiency can be achieved only when $K = M\cdot N$, the ratio of parameters can then be represented as $\frac{\frac{C}{M} + 4M + 9}{36C}$. Thus given $C$, $K$ and $F$, the best parameter efficiency can be reached by setting $\frac{C}{M} = 4M$, or $M = \frac{\sqrt{C}}{2}$.

\paragraph{Efficiency given the total amount of parameters.}

Similarly, according to the Theorem~\ref{the:eff} the greatest $C$ can be reached only when the number of bottleneck channels $K = M\cdot N$. Since $F = \alpha \cdot C$ and $K = F/4$, the total number of parameters of one design $P$ can be expressed as
\begin{align*}
    P & = 1\cdot 1\cdot \frac{4N}{\alpha}\cdot MN + 3\cdot 3\cdot MN + 1\cdot 1\cdot M\cdot 4MN = MN(\frac{4N}{\alpha}+9+4M) \\
    & \geq MN(9+2\sqrt{\frac{16MN}{\alpha}}) = \frac{\alpha}{4}C(9+4\sqrt{C})
\end{align*}
Given the number of parameters $P$, the greatest width $G$ exists when $\alpha M = N$.

\section{Connections to the State-of-the-Arts}

To verify the correctness of our scheme, we further explore the connections between designs in our scheme and other state-of-the-art ones.


\paragraph{GC + PWG.}

Xception and MobileNet are two popular efficient models with the same type of building block. After analyzing them carefully, one can find that they are just one extreme case of this design. Specifically, when $M$ is equal to the number of input channels and $N$ is 1 the design will become the building block of them.


\paragraph{PW + DW + PW.}

Our design can be regraded as an extreme case of ResNeXt when the number of groups is equal to the number of bottleneck channels. Later experiments in Section~\ref{sec:study} indicate that our design could achieve a better performance given the same number of parameters since more parameters could be saved for increasing the width of network.


\paragraph{PWG + DW + PWG.}

ShuffleNet has a similar building block with this design, both of which contain two pointwise group convolutions and a depthwise convolution. However, in ShuffleNet the two pointwise group convolutions share the same group number, whereas two distinct numbers could be used by our design, which could provide a better opportunity to reduce more parameters. Therefore ShuffleNet can be seen as a subset of our sparse kernel design. Experiments in Section~\ref{sec:sota} also show that given the same number of parameters our design could construct a wider model with better performance.


\section{Experiments}

We verify the idea of our scheme via experiments. First, we provide the implementation details for our experiments. Then we study the relationship between the~\emph{information field} and the final accuracy along with the comparisons between different SK designs in our scheme and the state-of-the-art ones. 

\subsection{Implementation Details}

\begin{table}[!th]
    \caption{Overall network layout. $B$ is the number of blocks at each stage. At the first block of each stage except the first stage down-sampling is performed and the channel number is doubled.}
    \label{tab:frame}
    \centering
    \begin{tabular}{lllll}
        \toprule
        Layer & Output size & KSize & Strides & Repeat \\
        \midrule
        Image & $224\times 224$ &  &  & \\
        \midrule
        Conv1 & $112\times 112$ & $3\times 3$ & 2 & 1\\
        \midrule
        Max Pool & $56\times 56$ & $3\times 3$ & 2 & 1\\
        Stage 1 & $56\times 56$ & & 1 & $B$\\
        \midrule
        Stage 2 & $28\times 28$ & & 2 & 1\\
        & $28\times 28$ & & 1 & $B-1$ \\
        \midrule
        Stage 3 & $14\times 14$ & & 2 & 1\\
        & $14\times 14$ & & 1 & $B-1$ \\
        \midrule
        Stage 4 & $7\times 7$ & & 2 & 1\\
        & $7\times 7$ & & 1 & $B-1$ \\
        \midrule
        Average Pool & $1\times 1$ & $7\times 7$ & & 1 \\
        \midrule
        \multicolumn{5}{c}{1000-d FC, Softmax} \\
        \bottomrule
    \end{tabular}
\end{table}

The overall layout of the network is shown in Table~\ref{tab:frame}. Identity mapping~\citep{he2016identity} is used over each block. When building the models, we can simply replace every block in the layout with the standard convolution or the SK designs mentioned in Section~\ref{sec:decompose}. Batch normalization (BN)~\citep{ioffe2015batch} is adopted right after each layer in the block and as suggested by~\citep{chollet2016xception} nonlinear activation ReLU is only performed after the summation of the identity shortcut and the output of each block. 

We evaluate our models on ImageNet 2012 dataset~\citep{deng2009imagenet,russakovsky2015imagenet}, which contains 1.2 million training images and 50000 validation images from 1000 categories. We follow the same data augmentation scheme in~\citep{he2016identity,he2016deep} which includes randomized cropping, color jittering and horizontal flipping. All models are trained for 100 epochs with batch size 256. SGD optimizer is used with the Nesterov momentum. The weight decay is 0.0001 and the momentum is 0.9. We adopt the similar weight initialization method from~\citep{he2015delving,he2016deep,huang2016densely}. The learning rate starts with 0.1 and is divided by 10 every 30 epochs. All results reported are single center crop top-1 performances.

\subsection{Empirical Study}
\label{sec:study}

\begin{table}[!th]
    \caption{Comparisons to illustrate the relationship between the~\emph{information field} and the model accuracy. We tune the number of group to achieve different parameter efficiency. Width here is the number of input channels to the first stage in the network. InfoSize is the size of~\emph{information field} with regards to the input to the first stage. Numbers within the parentheses represent the number of groups. For example, GConv(1) means group convolution with only 1 group, which is also the standard convolution.}
    \label{tab:width}
    \centering
    \begin{tabular}{llllll}
        \toprule
        Network Unit & \#Params($\times$M) & Depth & Width & InfoSize & Error (\%)\\
        \midrule
        PW+GConv(1)+PW & 13.9 & 98 & 128 & (3, 3, 128) & 30.0 \\
        PW+GConv(32)+PW & 13.9 & 98 & 256 & (3, 3, 256) & 29.2 \\
        \midrule
        PW+GConv(1)+PW & 28.4 & 194 & 128 & (3, 3, 128) & 29.7 \\
        PW+GConv(1)+PW & 28.4 & 98 & 200 & (3, 3, 200) & 29.3 \\
        \midrule
        PW+GConv(2)+PW & 28.4 & 98 & 256 & (3, 3, 256) & 28.7 \\
        PW+GConv(64)+PW & 28.4 & 98 & 512 & (3, 3, 512) & \textbf{28.4} \\
        \bottomrule
    \end{tabular}
\end{table}

\paragraph{Relationship between the~\emph{information field} and the model accuracy.}

In Section~\ref{sec:decompose}, we have shown that all the SK designs generated by our scheme share the same size of the~\emph{information field} when the size of input is fixed. Meanwhile different SK designs could save different amount of parameters/computation compared to the standard convolution and the saved computation/parameters can then be used to increase the number of channels, enlarge the~\emph{information field}, and increase the final accuracy. The fundamental idea behind this is that we believe the~\emph{information field} is an essential property of all SK designs and could directly affect the final accuracy.

To verify this idea we choose a bottleneck-like design and conduct some comparisons by tuning different number of groups. We adopt the same overall network layout in Table~\ref{tab:frame}. It can be easily verified that given the same size of the input tensor the change of the number of groups in the bottleneck-like design will not affect the size of the~\emph{information field} in the output. Results are shown in Table~\ref{tab:width}. Specifically, compare results on row 2 and row 5, we can see that by increasing the number of group from 2 to 32, more than a half amount of parameters will be saved to generate the same width, however the model accuracy will only decrease slightly. Meanwhile a further comparison on row 5 and row 6 indicate that if we use the saved parameters to increase the network width, the accuracy could still be improved. Since both of the two networks contain the same amount of parameters, overall network layout and type of SK design, the performance gains should only come from the increase of network width (~\emph{information field}). Same phenomenon could also be found by comparing results on row 1 and row 2. 

Besides we investigate on different usages of parameters, results on row 3 and row 4 show that the increase of network width has better potential for the improvement of accuracy than that of the depth, which also indicates that the size of the~\emph{information field} could play a more important role on the model accuracy. Additionally results in Table~\ref{tab:width} can further explain the SK design (PW+DW+PW) in Section~\ref{sec:PW+DW+PW} where we directly apply the most parameter-efficient depthwise convolution in the middle since it has the same size of the~\emph{information field} with other group numbers.

\paragraph{Comparisons of different SK designs.}

We also compare different SK designs mentioned in Section~\ref{sec:decompose}. Results are shown in Table~\ref{tab:decompose}. As mentioned in Section~\ref{sec:decompose} all designs have the same-sized~\emph{information field} given the same input. Results from Table~\ref{tab:decompose} show that given the close amount of parameters by choosing different SK designs or group numbers models with different widths can be constructed, and the final accuracy is positively correlated to the model width (the size of the~\emph{information field}), which also coincides with our analysis above. Also notice that results here do not necessarily indicate one type of SK design is always better than the other one in terms of the parameter efficiency since as per the analysis in Section~\ref{sec:decompose} the efficiency also depends on other factors like the number of groups. For example, considering the same number of parameters and overall network layout, there could be a combination of group numbers $M$ and $N$ such that the network with the design GConv($M$)+PWGConv($N$) is wider than that of DW+PW.

\begin{table}[!th]
    \caption{Comparisons of different SK designs. All designs share the same network layout.}
    \label{tab:decompose}
    \centering
    \begin{tabular}{lllll}
        \toprule
        Network Unit & \#Params($\times$M) & Width & InfoSize & Error (\%)\\
        \midrule
        Standard Convolution & 11.2 & 64 & (3, 3, 64) & 31.1  \\
        \midrule
        DW+PW & 0.8 & 72 & (3, 3, 72) &31.7 \\
        DW+PW & 11.2 & 280 & (3, 3, 280) & 28.5 \\
        \midrule
        GConv(4)+PWGConv(32) & 11.2 & 128 & (3, 3, 128) & 30.8 \\
        GConv(16)+PWGConv(16) & 11.3 & 256 & (3, 3, 256) & 29.4 \\
        \midrule
        PW+DW+PW & 11.0 & 400 & (3, 3, 400) &26.9\\
        \midrule
        PWGConv(4)+DW+PWGConv(4) & 11.3 & 560 & (3, 3, 560) & \textbf{25.6}\\
        \bottomrule
    \end{tabular}
\end{table}

\subsection{Comparisons with the State-of-the-Arts.}
\label{sec:sota}

\begin{table}[!th]
    \caption{Comparisons with different state-of-the-art SK designs. All settings are restored from the original papers. Specifically, bottleneck ratio is $1: 4$ for ResNet and ResNeXt adopts cardinality of 16 and bottleneck ratio of $1: 2$. Meanwhile 4 groups are used for ShuffleNet.}
    \label{tab:sota}
    \centering
    \begin{tabular}{lllll}
        \toprule
        Network Unit & \#Params($\times$M) & Width & InfoSize & Error (\%)\\
        \midrule
        ResNet~\citep{he2016deep} & 11.2 & 64 & (3, 3, 64) & 31.3 \\
        ResNet with bottleneck~\citep{he2016deep} & 11.3 & 192 & (3, 3, 192) &29.9 \\
        ResNeXt~\citep{xie2017aggregated} & 11.1 & 192 & (3, 3, 192) & 29.8 \\
        Xception~\citep{chollet2016xception} & 11.2 & 280 & (3, 3, 280) &28.5 \\
        ShuffleNet~\citep{zhang2017shufflenet} & 11.3 & 560 & (3, 3, 560) & 25.6\\
        \midrule
        GConv(100)+PWGConv(2) & 8.6 & 200 & (3, 3, 200) &27.0 \\
        PWGConv(100)+DW+PWGConv(2) & 10.4 & 700 & (3, 3, 700) & \textbf{24.9} \\
        \bottomrule
    \end{tabular}
\end{table}

Based on the SK scheme, we are also able to construct more efficient designs than the state-of-the-art ones. Table~\ref{tab:sota} shows comparisons between the SK designs generated by our scheme and the state-of-the-art ones. For fair comparisons, we use the same network layout as shown in Table~\ref{tab:frame} and replace blocks in it with corresponding designs, and the model size around 11.0M is selected as it is the size that different models (e.g., Xception, ResNeXt and ShuffleNet) can be easily configured to. Results in Table~\ref{tab:sota} indicate that SK designs in our scheme could even yield better accuracy with a smaller model size, which also validates the idea of our SK scheme. Also notice that the choices of group numbers used in our designs are chosen to help easily accommodate both the similar model size and the overall network layout, which may not be the most efficient ones that are supposed to result in a wider network with better accuracy under the same limitation of parameters.

\section{Related Works}




\paragraph{Model Compression.}

Traditional model compression techniques include pruning, quantization and low-rank approximation. Pruning~\citep{wen2016learning,ardakani2016sparsely,liu2017learning,li2016pruning,he2017channel,liu2015sparse} reduces redundant weights, network connections or channels in a pre-trained model. However, it could face difficulty for deploying on hardware like GPU since some pruning methods may be only effective when the weight matrix is sufficiently sparse. Quantization~\citep{zhou2016dorefa,zhou2017incremental,courbariaux2015binaryconnect,courbariaux2016binarized,deng2018gxnor,micikevicius2017mixed} reduces the number of bits required to represent weights. Unfortunately, this technique will require specialized hardware support. Low rank approximation~\citep{lebedev2014speeding,jin2014flattened,wang2016design,xue2014singular,novikov2015tensorizing,garipov2016ultimate} uses two or more matrices to approximate the original matrix values in a pre-trained model. Nevertheless, since the process is an approximation of original matrix values maintaining a similar accuracy will always need additional re-training. The focus of this paper, the SK approach, mitigates all these problems by directly training networks using structural sparse convolutional kernels.


\section{Conclusion}

In this paper, we present a scheme to craft the effective SK design by eliminating the large design space from three aspects: composition, performance and efficiency. During the process to reduce the design space, we find an unified property named~\emph{information field} behind various designs, which could directly indicate the final accuracy. Meanwhile we show the final 4 designs in our scheme along with detailed efficiency analysis. Experimental results also validate the idea of our scheme.

\bibliography{main}

\begin{thebibliography}{36}
\providecommand{\natexlab}[1]{#1}
\providecommand{\url}[1]{\texttt{#1}}
\expandafter\ifx\csname urlstyle\endcsname\relax
  \providecommand{\doi}[1]{doi: #1}\else
  \providecommand{\doi}{doi: \begingroup \urlstyle{rm}\Url}\fi

\bibitem[Ardakani et~al.(2016)Ardakani, Condo, and Gross]{ardakani2016sparsely}
Arash Ardakani, Carlo Condo, and Warren~J Gross.
\newblock Sparsely-connected neural networks: towards efficient vlsi
  implementation of deep neural networks.
\newblock \emph{arXiv preprint arXiv:1611.01427}, 2016.

\bibitem[Chollet(2016)]{chollet2016xception}
Fran{\c{c}}ois Chollet.
\newblock Xception: Deep learning with depthwise separable convolutions.
\newblock \emph{arXiv preprint}, 2016.

\bibitem[Courbariaux et~al.(2015)Courbariaux, Bengio, and
  David]{courbariaux2015binaryconnect}
Matthieu Courbariaux, Yoshua Bengio, and Jean-Pierre David.
\newblock Binaryconnect: Training deep neural networks with binary weights
  during propagations.
\newblock In \emph{Advances in neural information processing systems}, pp.\
  3123--3131, 2015.

\bibitem[Courbariaux et~al.(2016)Courbariaux, Hubara, Soudry, El-Yaniv, and
  Bengio]{courbariaux2016binarized}
Matthieu Courbariaux, Itay Hubara, Daniel Soudry, Ran El-Yaniv, and Yoshua
  Bengio.
\newblock Binarized neural networks: Training deep neural networks with weights
  and activations constrained to+ 1 or-1.
\newblock \emph{arXiv preprint arXiv:1602.02830}, 2016.

\bibitem[Deng et~al.(2009)Deng, Dong, Socher, Li, Li, and
  Fei-Fei]{deng2009imagenet}
Jia Deng, Wei Dong, Richard Socher, Li-Jia Li, Kai Li, and Li~Fei-Fei.
\newblock Imagenet: A large-scale hierarchical image database.
\newblock In \emph{Computer Vision and Pattern Recognition, 2009. CVPR 2009.
  IEEE Conference on}, pp.\  248--255. IEEE, 2009.

\bibitem[Deng et~al.(2018)Deng, Jiao, Pei, Wu, and Li]{deng2018gxnor}
Lei Deng, Peng Jiao, Jing Pei, Zhenzhi Wu, and Guoqi Li.
\newblock Gxnor-net: Training deep neural networks with ternary weights and
  activations without full-precision memory under a unified discretization
  framework.
\newblock \emph{Neural Networks}, 100:\penalty0 49--58, 2018.

\bibitem[Denton et~al.(2014)Denton, Zaremba, Bruna, LeCun, and
  Fergus]{denton2014exploiting}
Emily~L Denton, Wojciech Zaremba, Joan Bruna, Yann LeCun, and Rob Fergus.
\newblock Exploiting linear structure within convolutional networks for
  efficient evaluation.
\newblock In \emph{Advances in neural information processing systems}, pp.\
  1269--1277, 2014.

\bibitem[Garipov et~al.(2016)Garipov, Podoprikhin, Novikov, and
  Vetrov]{garipov2016ultimate}
Timur Garipov, Dmitry Podoprikhin, Alexander Novikov, and Dmitry Vetrov.
\newblock Ultimate tensorization: compressing convolutional and fc layers
  alike.
\newblock \emph{arXiv preprint arXiv:1611.03214}, 2016.

\bibitem[He et~al.(2015)He, Zhang, Ren, and Sun]{he2015delving}
Kaiming He, Xiangyu Zhang, Shaoqing Ren, and Jian Sun.
\newblock Delving deep into rectifiers: Surpassing human-level performance on
  imagenet classification.
\newblock In \emph{Proceedings of the IEEE international conference on computer
  vision}, pp.\  1026--1034, 2015.

\bibitem[He et~al.(2016{\natexlab{a}})He, Zhang, Ren, and Sun]{he2016deep}
Kaiming He, Xiangyu Zhang, Shaoqing Ren, and Jian Sun.
\newblock Deep residual learning for image recognition.
\newblock In \emph{Proceedings of the IEEE conference on computer vision and
  pattern recognition}, pp.\  770--778, 2016{\natexlab{a}}.

\bibitem[He et~al.(2016{\natexlab{b}})He, Zhang, Ren, and Sun]{he2016identity}
Kaiming He, Xiangyu Zhang, Shaoqing Ren, and Jian Sun.
\newblock Identity mappings in deep residual networks.
\newblock In \emph{European Conference on Computer Vision}, pp.\  630--645.
  Springer, 2016{\natexlab{b}}.

\bibitem[He et~al.(2017)He, Zhang, and Sun]{he2017channel}
Yihui He, Xiangyu Zhang, and Jian Sun.
\newblock Channel pruning for accelerating very deep neural networks.
\newblock In \emph{International Conference on Computer Vision (ICCV)},
  volume~2, 2017.

\bibitem[Howard et~al.(2017)Howard, Zhu, Chen, Kalenichenko, Wang, Weyand,
  Andreetto, and Adam]{howard2017mobilenets}
Andrew~G Howard, Menglong Zhu, Bo~Chen, Dmitry Kalenichenko, Weijun Wang,
  Tobias Weyand, Marco Andreetto, and Hartwig Adam.
\newblock Mobilenets: Efficient convolutional neural networks for mobile vision
  applications.
\newblock \emph{arXiv preprint arXiv:1704.04861}, 2017.

\bibitem[Huang et~al.(2016)Huang, Liu, Weinberger, and van~der
  Maaten]{huang2016densely}
Gao Huang, Zhuang Liu, Kilian~Q Weinberger, and Laurens van~der Maaten.
\newblock Densely connected convolutional networks.
\newblock \emph{arXiv preprint arXiv:1608.06993}, 2016.

\bibitem[Ioffe \& Szegedy(2015)Ioffe and Szegedy]{ioffe2015batch}
Sergey Ioffe and Christian Szegedy.
\newblock Batch normalization: Accelerating deep network training by reducing
  internal covariate shift.
\newblock \emph{arXiv preprint arXiv:1502.03167}, 2015.

\bibitem[Jin et~al.(2014)Jin, Dundar, and Culurciello]{jin2014flattened}
Jonghoon Jin, Aysegul Dundar, and Eugenio Culurciello.
\newblock Flattened convolutional neural networks for feedforward acceleration.
\newblock \emph{arXiv preprint arXiv:1412.5474}, 2014.

\bibitem[Krizhevsky et~al.(2012)Krizhevsky, Sutskever, and
  Hinton]{krizhevsky2012imagenet}
Alex Krizhevsky, Ilya Sutskever, and Geoffrey~E Hinton.
\newblock Imagenet classification with deep convolutional neural networks.
\newblock In \emph{Advances in neural information processing systems}, pp.\
  1097--1105, 2012.

\bibitem[Lebedev et~al.(2014)Lebedev, Ganin, Rakhuba, Oseledets, and
  Lempitsky]{lebedev2014speeding}
Vadim Lebedev, Yaroslav Ganin, Maksim Rakhuba, Ivan Oseledets, and Victor
  Lempitsky.
\newblock Speeding-up convolutional neural networks using fine-tuned
  cp-decomposition.
\newblock \emph{arXiv preprint arXiv:1412.6553}, 2014.

\bibitem[LeCun et~al.(1990)LeCun, Denker, and Solla]{lecun1990optimal}
Yann LeCun, John~S Denker, and Sara~A Solla.
\newblock Optimal brain damage.
\newblock In \emph{Advances in neural information processing systems}, pp.\
  598--605, 1990.

\bibitem[Li et~al.(2016)Li, Kadav, Durdanovic, Samet, and Graf]{li2016pruning}
Hao Li, Asim Kadav, Igor Durdanovic, Hanan Samet, and Hans~Peter Graf.
\newblock Pruning filters for efficient convnets.
\newblock \emph{arXiv preprint arXiv:1608.08710}, 2016.

\bibitem[Lin et~al.(2013)Lin, Chen, and Yan]{lin2013network}
Min Lin, Qiang Chen, and Shuicheng Yan.
\newblock Network in network.
\newblock \emph{arXiv preprint arXiv:1312.4400}, 2013.

\bibitem[Liu et~al.(2015)Liu, Wang, Foroosh, Tappen, and Pensky]{liu2015sparse}
Baoyuan Liu, Min Wang, Hassan Foroosh, Marshall Tappen, and Marianna Pensky.
\newblock Sparse convolutional neural networks.
\newblock In \emph{Proceedings of the IEEE Conference on Computer Vision and
  Pattern Recognition}, pp.\  806--814, 2015.

\bibitem[Liu et~al.(2017)Liu, Li, Shen, Huang, Yan, and Zhang]{liu2017learning}
Zhuang Liu, Jianguo Li, Zhiqiang Shen, Gao Huang, Shoumeng Yan, and Changshui
  Zhang.
\newblock Learning efficient convolutional networks through network slimming.
\newblock In \emph{Computer Vision (ICCV), 2017 IEEE International Conference
  on}, pp.\  2755--2763. IEEE, 2017.

\bibitem[Micikevicius et~al.(2017)Micikevicius, Narang, Alben, Diamos, Elsen,
  Garcia, Ginsburg, Houston, Kuchaev, Venkatesh, et~al.]{micikevicius2017mixed}
Paulius Micikevicius, Sharan Narang, Jonah Alben, Gregory Diamos, Erich Elsen,
  David Garcia, Boris Ginsburg, Michael Houston, Oleksii Kuchaev, Ganesh
  Venkatesh, et~al.
\newblock Mixed precision training.
\newblock \emph{arXiv preprint arXiv:1710.03740}, 2017.

\bibitem[Novikov et~al.(2015)Novikov, Podoprikhin, Osokin, and
  Vetrov]{novikov2015tensorizing}
Alexander Novikov, Dmitrii Podoprikhin, Anton Osokin, and Dmitry~P Vetrov.
\newblock Tensorizing neural networks.
\newblock In \emph{Advances in Neural Information Processing Systems}, pp.\
  442--450, 2015.

\bibitem[Russakovsky et~al.(2015)Russakovsky, Deng, Su, Krause, Satheesh, Ma,
  Huang, Karpathy, Khosla, Bernstein, et~al.]{russakovsky2015imagenet}
Olga Russakovsky, Jia Deng, Hao Su, Jonathan Krause, Sanjeev Satheesh, Sean Ma,
  Zhiheng Huang, Andrej Karpathy, Aditya Khosla, Michael Bernstein, et~al.
\newblock Imagenet large scale visual recognition challenge.
\newblock \emph{International Journal of Computer Vision}, 115\penalty0
  (3):\penalty0 211--252, 2015.

\bibitem[Sifre \& Mallat(2014)Sifre and Mallat]{sifre2014rigid}
Laurent Sifre and St{\'e}phane Mallat.
\newblock Rigid-motion scattering for image classification.
\newblock 2014.

\bibitem[Simonyan \& Zisserman(2014)Simonyan and Zisserman]{simonyan2014very}
Karen Simonyan and Andrew Zisserman.
\newblock Very deep convolutional networks for large-scale image recognition.
\newblock \emph{arXiv preprint arXiv:1409.1556}, 2014.

\bibitem[Soudry et~al.(2014)Soudry, Hubara, and Meir]{soudry2014expectation}
Daniel Soudry, Itay Hubara, and Ron Meir.
\newblock Expectation backpropagation: Parameter-free training of multilayer
  neural networks with continuous or discrete weights.
\newblock In \emph{Advances in Neural Information Processing Systems}, pp.\
  963--971, 2014.

\bibitem[Wang et~al.(2016)Wang, Liu, and Foroosh]{wang2016design}
Min Wang, Baoyuan Liu, and Hassan Foroosh.
\newblock Design of efficient convolutional layers using single intra-channel
  convolution, topological subdivisioning and spatial" bottleneck" structure.
\newblock \emph{arXiv preprint arXiv:1608.04337}, 2016.

\bibitem[Wen et~al.(2016)Wen, Wu, Wang, Chen, and Li]{wen2016learning}
Wei Wen, Chunpeng Wu, Yandan Wang, Yiran Chen, and Hai Li.
\newblock Learning structured sparsity in deep neural networks.
\newblock In \emph{Advances in Neural Information Processing Systems}, pp.\
  2074--2082, 2016.

\bibitem[Xie et~al.(2017)Xie, Girshick, Doll{\'a}r, Tu, and
  He]{xie2017aggregated}
Saining Xie, Ross Girshick, Piotr Doll{\'a}r, Zhuowen Tu, and Kaiming He.
\newblock Aggregated residual transformations for deep neural networks.
\newblock In \emph{Computer Vision and Pattern Recognition (CVPR), 2017 IEEE
  Conference on}, pp.\  5987--5995. IEEE, 2017.

\bibitem[Xue et~al.(2014)Xue, Li, Yu, Seltzer, and Gong]{xue2014singular}
Jian Xue, Jinyu Li, Dong Yu, Mike Seltzer, and Yifan Gong.
\newblock Singular value decomposition based low-footprint speaker adaptation
  and personalization for deep neural network.
\newblock In \emph{Acoustics, Speech and Signal Processing (ICASSP), 2014 IEEE
  International Conference on}, pp.\  6359--6363. IEEE, 2014.

\bibitem[Zhang et~al.(2017)Zhang, Zhou, Lin, and Sun]{zhang2017shufflenet}
Xiangyu Zhang, Xinyu Zhou, Mengxiao Lin, and Jian Sun.
\newblock Shufflenet: An extremely efficient convolutional neural network for
  mobile devices.
\newblock \emph{arXiv preprint arXiv:1707.01083}, 2017.

\bibitem[Zhou et~al.(2017)Zhou, Yao, Guo, Xu, and Chen]{zhou2017incremental}
Aojun Zhou, Anbang Yao, Yiwen Guo, Lin Xu, and Yurong Chen.
\newblock Incremental network quantization: Towards lossless cnns with
  low-precision weights.
\newblock \emph{arXiv preprint arXiv:1702.03044}, 2017.

\bibitem[Zhou et~al.(2016)Zhou, Wu, Ni, Zhou, Wen, and Zou]{zhou2016dorefa}
Shuchang Zhou, Yuxin Wu, Zekun Ni, Xinyu Zhou, He~Wen, and Yuheng Zou.
\newblock Dorefa-net: Training low bitwidth convolutional neural networks with
  low bitwidth gradients.
\newblock \emph{arXiv preprint arXiv:1606.06160}, 2016.

\end{thebibliography}
\bibliographystyle{main}

\section{Appendix}
\textbf{Proof of Theorem~\ref{the:eff}}
\begin{proof}
Without loss of generality we use the example in Section~\ref{sec:proof} to prove the theorem. Recall that the total number of parameters for such design can be expressed as
\begin{equation*}
    P = 3\cdot 3\cdot \frac{C}{M}\cdot C + 1\cdot 1\cdot \frac{C}{N}\cdot F = \frac{9C^2}{M} + \frac{CF}{N},
\end{equation*}
then the problem could be interpreted as proving that the minimum value of $P$ can be achieved if and only if $M\cdot N = C$.

We prove the theorem by contradiction. Assume the minimum value of $P$ could be achieved when $M\cdot N < C$. Then we can always find a $N' = C/M > N$ such that the combination of $M$ and $N'$ could result in a smaller value of $P$, which contradicts our assumption. The theorem is hence proved.
\end{proof}
\end{spacing}
\end{document}